# Putting GPT-4o to the Sword: A Comprehensive Evaluation of Language, Vision, Speech, and Multimodal Proficiency

Sakib Shahriar[1], Brady Lund[2], Nishith Reddy Mannuru[2], Muhammad Arbab Arshad[3], Kadhim Hayawi[4], Ravi Varma Kumar Bevara[2], Aashrith Mannuru[5], Laiba Batool[6]

[1]University of Guelph, Guelph, ON, Canada

[2]University of North Texas, Denton, TX, USA

[3]Iowa State University, Ames, IA, USA

[4]College of Interdisciplinary Studies, Zayed University, United Arab Emirates

[5]University of Texas at Dallas, Richardson, TX, USA

[6]National University of Computer & Emerging Sciences, Islamabad, Pakistan

**Abstract**

As large language models (LLMs) continue to advance, evaluating their comprehensive capabilities becomes significant for their application in various fields. This research study comprehensively evaluates the language, vision, speech, and multimodal capabilities of GPT-4o. The study employs standardized exam questions, reasoning tasks, and translation assessments to assess the model's language capability. Additionally, GPT-4o's vision and speech capabilities are tested through image classification and object recognition tasks, as well as accent classification. The multimodal evaluation assesses the model's performance in integrating visual and linguistic data. Our findings reveal that GPT-4o demonstrates high accuracy and efficiency across multiple domains in language and reasoning capabilities, excelling in tasks that require few-shot learning. GPT-4o also provides notable improvements in multimodal tasks compared to its predecessors. However, the model shows variability and faces limitations in handling complex and ambiguous inputs, particularly in audio and vision capabilities. This paper highlights the need for more comprehensive benchmarks and robust evaluation frameworks, encompassing qualitative assessments involving human judgment as well as error analysis. Future work should focus on expanding datasets, investigating prompt-based assessment, and enhancing few-shot learning techniques to test the model's practical applicability and performance in real-world scenarios.

# 1. Introduction

In the past few years, the emergence of large language models has led to paradigm shifts across various disciplines and professions. The pursuit of building and implementing the most powerful and accurate models has captured both researchers and industry. In late 2023 and early 2024, competitors to OpenAI, including Google and Anthropic, introduced advanced large language models: Google's Gemini and Anthropic's Claude 3 (Gemini Team et al., 2024; Korinek, 2023). These models surpassed the capabilities of the original GPT-3, GPT-3.5, and GPT-4 models that powered ChatGPT. To stay competitive, OpenAI needed to develop an upgraded model with more parameters, enhanced capabilities, and improved speed. This led to the launch of GPT-4 Omni (GPT-4o) in May 2024.

GPT-4o introduces several major innovations that improve upon previous large language models. The model includes a massive number of parameters – estimated to be well over one trillion – which dwarfs GPT-3, at 175 billion parameters, and GPT-1, at an estimated 117 million parameters (Floridi & Chiriatti, 2020). The model is able to process and generate text, image, and audio content and does so at a speed that is much faster than competitor models. Importantly, the model also integrates improved handling of ambiguous and complex queries, where a misunderstanding could emerge between the user and the model, and enhances its ethical and safety protocols to mitigate the prevalence of harmful or incorrect outputs, as has been an issue with competitor models in recent months (Dillion et al., 2024; Ray, 2024). Though all these innovations appear to be a tremendous boon for the model, there are many areas where the efficacy of the model has not yet been formally evaluated.

## 1.1 Research Purpose

The purpose of this study is to comprehensively evaluate the capabilities of GPT-4 Omni (GPT-4o) across various domains, including language, vision, speech, and multimodal tasks. By systematically assessing GPT-4o's performance on a wide range of benchmarks and real-world tasks, we aim to understand its capabilities, strengths, and limitations. This evaluation will provide insights into the advancements made by GPT-4o compared to previous models, such as GPT-3 and GPT-4, and other contemporary models like Google's Gemini and Anthropic's Claude 3. These findings will contribute to ongoing investigations of the practical applications and future development of large language models.

## 1.2 Related Work

GPT-4o is the latest development in a string of innovations to generative pre-trained transformers in recent years. In order to situate the development of GPT-4o within the context of the greater developments occurring in artificial intelligence (AI), it may be helpful to view these technologies as a series of nested boxes, as in Figure 1. AI as a concept encompasses a wide range of developments, of which machine learning and deep learning are but one area (Ongsulee, 2017). Within deep learning, there are further divisions, with generative AI being only one (albeit major) area. The same is true for large language models, as one application of generative AI. We already know of other types of generative AI that are not language-based, such as image generators. The generative pre-trained transformer is but one large language model (LLM), developed by OpenAI. GPT-4o is the latest version of this model. As such, while GPT-4o is a very important innovation, it is but one element within the broad AI landscape that exists today.

**Figure 1. Visualization of the Relationship Between General AI and GPT 4o**

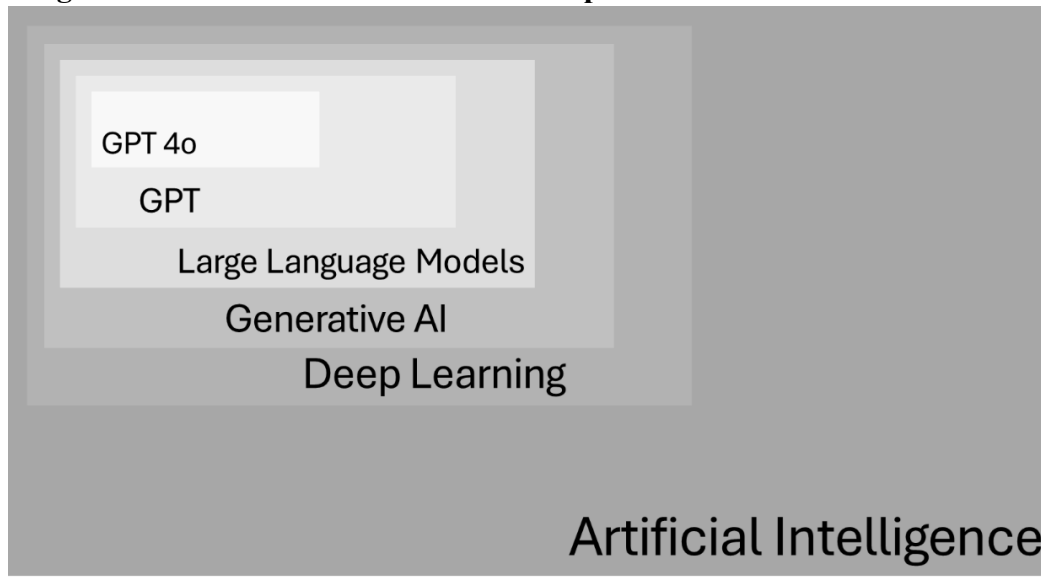

As illustrated in Figure 1, GPT-4o belongs to the class of technologies known as large language models (LLMs). These models are notable for their ability to mimic human language usage so closely that it can be difficult for a human observer to distinguish between text generated by a human and that generated by a machine (Thirunavukarasu et al., 2023; Hayawi et al., 2024). This innovation marks a significant advancement towards passing the Turing test and underscores the practicality of AI in writing and research (Aher et al., 2023; Mannuru et al., 2023). However, it also introduces significant risks, including potential invasions of privacy and the generation of inaccurate, misleading, biased, or harmful information (Lund et al., 2023). Therefore, it is crucial to carefully evaluate these LLMs and scrutinize their outputs. Failure to do so could lead to the proliferation of misinformation and malicious content on the Internet (Hu et al., 2024).

Given the serious issues associated with some LLMs, it is essential to critically examine each new model for its limitations. Recent versions of GPT have shown significant improvements over their predecessors in various areas. For example, Koubaa (2023) found substantial improvements in GPT-4 compared to GPT-3.5 on tests such as the Graduate Record Examination (GRE), SAT, and Bar exam, with GPT-4's performance placing it in the top tenth percentile on most of these exams. Similarly, Coyne et al. (2023) reported improvements in grammatical error correction for GPT-4 compared to GPT-3.5. However, having more parameters in a model does not inherently guarantee better performance on all tasks. Overfitting can occur when a model is extensively trained on a large dataset but fails to generalize well to real-world data (Salman & Liu, 2019).

Evaluation of the GPT-4o model is currently very limited. Research has explored various aspects of the model, including potential threats (Shen et al., 2024; Ying et al., 2024), diagnostic ability (Oura et al., 2024; Zhang et al., n.d.), and multilingual capabilities (Wang et al., 2024). One study by Sonoda et al. (2024) found that GPT-4o underperforms compared to Claude 3 Opus in radiology diagnosis tasks. Other studies investigated the sentiment of the general public regarding GPT-4o (Singgalen, 2024). Many studies focus on ChatGPT, the chatbot powered by GPT models, rather than the models themselves. These studies provide some additional insights into the quality of the models, such as their performance in English language teaching tasks (Pang et al., 2024). However, a comprehensive evaluation of GPT-4o itself remains a gap in the literature.

GPT-4o lends itself to new forms of evaluation beyond the language and reasoning evaluation of past model versions due to its new capabilities in vision, speech, and cross-modal activities. GPT-4 with Vision (GPT-4V) was previously evaluated on vision tasks; however, it is clear from these studies that the model was not ready for the visual challenges to which it was exposed (Xu et al., 2024; Zhou et al., 2024). Meanwhile, the speech capacity of GPT-4o is a new innovation, one that has already been met with some criticism due to the choices for the voice of the model (Allyn, 2024). While cross-modal activities have been theorized in LLMs for some time, GPT-4V stands out as among the first models to actualize this potential, paving the way for its evaluation (Li et al., 2021).

## 2. Language Capacity of GPT-4o

Language capacity is foundational to developing intelligent models capable of understanding, generating, and interacting with human language. This capacity encompasses a range of skills that enable models to process and produce coherent and contextually appropriate responses in natural language. The objective of this section is to comprehensively assess the language performance of GPT-4o (omni) by testing it on exams, reasoning tasks, and translation activities. Each of these tasks is significant for evaluating different aspects of the model's language capabilities.

### 2.1 Performance on Exams

In this subsection, we evaluate GPT-4o's performance on various standardized and board exam questions. This helps us gauge the model's ability to comprehend complex problems and generate coherent, relevant, and accurate responses. Standardized exams are designed to measure a range of cognitive abilities and knowledge across different subjects. This task measures the model's proficiency in handling structured questions across various subjects. Our methods involve presenting GPT-4o with questions from a variety of standardized and board exams. The responses generated by GPT-4o are then analyzed based on the correctness of the answers provided.

**Performance on USMLE**

The United States Medical Licensing Examination (USMLE) Step 1 is a rigorous and comprehensive assessment designed to evaluate a candidate's understanding and ability to apply key concepts in medical science necessary for the practice of medicine (Federation of State Medical Boards & of Medical Examiners, 2024). Jointly developed by the Federation of State Medical Boards and the National Board of Medical Examiners, this examination serves as a milestone for medical students and professionals aiming to obtain their medical licensure in the United States. The USMLE Step 1 primarily focuses on testing the examinee's grasp of foundational medical knowledge and their ability to apply this knowledge to clinical scenarios. The sample test questions provided in the USMLE Step 1 Sample Items booklet encompass various disciplines, including anatomy, biochemistry, microbiology, pathology, pharmacology, physiology, and interdisciplinary areas such as genetics, immunology, and molecular biology. The dataset used for evaluating GPT-4o's performance includes 119 sample test questions from the USMLE Step 1 booklet, updated as of January 2024[1].

Out of the total 118 questions, GPT-4o correctly answered 98 questions. This corresponds to an accuracy of 83.1%. Table 1 provides a comparison of GPT-4o with its predecessor models, as reported by Gilson et al. (2023) and Brin et al. (2023). Compared to its predecessor, GPT-3.5, which achieved an accuracy of 51.67%, GPT-4o shows significant improvement. GPT-4o, despite being designed for faster and more efficient tasks, offers a notable enhancement in language comprehension and problem-solving capabilities. However, GPT-4o's performance is slightly lower than that of GPT-4, which achieved an accuracy of 90.00%. This decline can be attributed to the design focus of GPT-4o on efficiency and speed, while GPT-4 remains the model for more complex and demanding tasks.

---

[1] https://www.usmle.org/sites/default/files/2021-10/Step_1_Sample_Items.pdf

The results indicate that GPT-4o can serve as a valuable tool in medical education, offering fast, interactive learning experiences that are crucial for students needing immediate feedback and guidance (Haleem et al., 2022). While GPT-4 excels in handling more intricate questions, its slower response time may limit its practicality for real-time learning scenarios. Meanwhile, GPT-4o's accuracy and efficiency make it suitable for dynamic educational environments.

| Model   | Total Questions | Correct Answers | Accuracy |
| ------- | --------------- | --------------- | -------- |
| GPT-3.5 | 389             | 201             | 51.67%   |
| GPT-4   | 80              | 72              | 90.00%   |
| GPT-4o  | 118             | 98              | 83.05%   |

Table 1. Performance Comparison of GPT models on USMLE

**Performance on CFA**

The Chartered Financial Analyst (CFA) Level 1 exam is a globally recognized certification offered by the CFA Institute, aimed at financial and investment professionals (CFA Institute, n.d.). The exam covers a broad range of topics, including ethical and professional standards, quantitative methods, economics, corporate finance, equity investments, fixed income, derivatives, and portfolio management. The CFA Level 1 exam is known for its rigorous and comprehensive assessment of a candidate's foundational knowledge and skills in finance and investment. It tests both theoretical understanding and the practical application of financial concepts and principles.

For this evaluation, we utilized the dataset from the 300Hours CFA Level 1 Mock Exam, which includes questions developed to mirror the style and difficulty of the actual exam[2]. GPT-4o correctly answered 76 out of the 89 questions, yielding an overall accuracy of 85.39%. Table 2 summarizes the performance in comparison to GPT-3.5 and GPT-4, as reported by Callanan et al. (2023). We compare the results obtained using zero-shot prompting since we did not provide the models with any hints or specific instructions during our prompting. The results indicate that GPT-4o noticeably outperforms both its predecessors. The increased accuracy of GPT-4o (despite being designed for faster and more efficient tasks) indicates that it can provide reliable and timely assistance for financial exam preparation.

| Model   | Accuracy |
| ------- | -------- |
| GPT-3.5 | 58.80%   |
| GPT-4   | 73.20%   |
| GPT-4o  | 85.39%   |

Table 2. Performance Comparison of GPT models on CFA Level 1 Exam

---

[2] https://300hours.com/free-cfa-level-1-mock-exam/

**Performance on SAT**

The Scholastic Assessment Test (SAT) is a standardized test widely used for college admissions in the United States (College Board, n.d.). Developed and administered by the College Board, the SAT assesses a student's readiness for college and provides colleges with a common data point for comparing all applicants. The SAT covers areas including reading, writing and language, and mathematics, with an optional essay section. This test is designed to measure a range of skills necessary for academic success in college, including critical thinking, problem-solving, and analytical abilities.

The dataset used for evaluating GPT-4o's performance consists of questions from the SAT Practice Test #1, which includes a variety of reading, writing, and math questions that reflect the format and content of the actual SAT exam[3]. The practice test consisted of two modules, each containing a reading and writing exam as well as a math exam. The performance on each module are outlined in Table 3.

| Test | Total Questions | Correct Answers | Accuracy |
|---|---|---|---|
| Reading & Writing M1 | 33 | 31 | 93.94% |
| Reading & Writing M2 | 33 | 29 | 87.88% |
| Math M1 | 27 | 25 | 92.59% |
| Math M2 | 27 | 22 | 81.48% |

Table 3. GPT-4o Performance on SAT

For comparison with previous GPT models, we refer to the comprehensive report by the Open AI team (OpenAI et al., 2024). In this context, we average the results of M1 and M2 for GPT-4o, as summarized in Table 4.

| Model | Reading & Writing | Math |
|---|---|---|
| GPT-3.5 | 83.75% | 73.75% |
| GPT-4 | 88.75% | 87.50% |
| GPT-4 (no vision) | 88.75% | 86.25% |
| GPT-4o | 90.91% | 87.04% |

Table 4. Performance Comparison of GPT models on SAT

GPT-4o demonstrates the highest accuracy in the Reading & Writing section with 90.91%, surpassing all the older models. In the Math section, GPT-4o achieves a strong performance with 87.04%, slightly lower than GPT-4 but higher than the rest. Figures 2-5 provide examples of GPT-4o correct and incorrect responses on each of the SAT categories.

---

[3] https://satsuite.collegeboard.org/sat/practice-preparation?excmpid=mtg796-st-1-bk

**Figure 2. Example of GPT-4o Incorrect Answer on SAT Reading & Writing: Question (Top), GPT-4o Response (Bottom Left), and Correct Answer (Bottom Right)**

| | |
|---|---|
| Lucía Michel of the University of Chile observed that alkaline soils contain an insoluble form of iron that blueberry plants cannot absorb, thus inhibiting blueberry growth. If these plants were grown in alkaline soil alongside grasses that aid in iron solubilization, _______ Michel was determined to find out. | Which choice completes the text so that it conforms to the conventions of Standard English?<br>A) could the blueberries thrive.<br>B) the blueberries could thrive.<br>C) the blueberries could thrive?<br>D) could the blueberries thrive? |
| B - The blueberries could thrive.<br><br>• Explanation: This phrasing correctly completes the sentence as a conditional statement. | **Choice D** is the best answer. The convention being tested is end-of-sentence punctuation. This choice correctly uses a question mark to punctuate the interrogative clause "could the blueberries thrive," which asks a direct question at the end of the sentence. |

**Figure 3. Example of GPT-4o Correct Answer on SAT Reading & Writing: Question (Top), GPT-4o Response (Bottom Left), and Correct Answer (Bottom Right)**

| | |
|---|---|
| While researching a topic, a student has taken the following notes:<br>• In the midst of the US Civil War, Susie Taylor escaped slavery and fled to Union-army-occupied St. Simons Island off the Georgia coast.<br>• She began working for an all-Black army regiment as a nurse and teacher.<br>• In 1902, she published a book about the time she spent with the regiment.<br>• Her book was the only Civil War memoir to be published by a Black woman.<br>• It is still available to readers in print and online. | The student wants to emphasize the uniqueness of Taylor's accomplishment. Which choice most effectively uses relevant information from the notes to accomplish this goal?<br>A) Taylor fled to St. Simons Island, which was then occupied by the Union army, for whom she began working.<br>B) After escaping slavery, Taylor began working for an all-Black army regiment as a nurse and teacher.<br>C) The book Taylor wrote about the time she spent with the regiment is still available to readers in print and online.<br>D) Taylor was the only Black woman to publish a Civil War memoir. |
| D - Taylor was the only Black woman to publish a Civil War memoir.<br><br>• Explanation: This option emphasizes the uniqueness of Taylor's accomplishment. | **Choice D** is the best answer. By indicating that Taylor's book was the only Civil War memoir published by a Black woman, this sentence emphasizes the uniqueness, or one-of-a-kind nature, of Taylor's accomplishment. |

**Figure 4. Example of GPT-4o Incorrect Answer on SAT Math: Question (Top), GPT-4o Response (Bottom Left), and Correct Answer (Bottom Right)**

| | |
|---|---|
| [Graph showing a line from (0, 40) to (60, 0) with y-axis labeled "Company B" from 0 to 50 and x-axis labeled "Company A" from 0 to 100] | The graph shows the relationship between the number of shares of stock from Company A, $x$, and the number of shares of stock from Company B, $y$, that Simone can purchase. Which equation could represent this relationship?<br>A) $y = 8x + 12$<br>B) $8x + 12y = 480$<br>C) $y = 12x + 8$<br>D) $12x + 8y = 480$ |

**Solution:**

Observing the graph, the line has a negative slope, which means as $x$ increases, $y$ decreases. The equation that matches this relationship is in the form of a linear equation that, when solved for $y$, has a negative coefficient for $x$.

By rearranging and comparing:
$8x + 12y = 480 \implies y = \frac{480-8x}{12} = 40 - \frac{2}{3}x$
$12x + 8y = 480 \implies y = \frac{480-12x}{8} = 60 - 1.5x$

The second equation matches the linear decrease observed in the graph.

**Answer:** D) 12x + 8y = 480

---

**Choice B** is correct. The graph shown is a line passing through the points $(0,40)$ and $(60,0)$. Since the relationship between $x$ and $y$ is linear, if two points on the graph make a linear equation true, then the equation represents the relationship. Substituting 0 for $x$ and 40 for $y$ in the equation in choice B, $8x+12y=480$, yields $8(0)+12(40)=480$, or $480=480$, which is true. Substituting 60 for $x$ and 0 for $y$ in the equation $8x+12y=480$ yields $8(60)+12(0)=480$, or $480=480$, which is true. Therefore, the equation $8x+12y=480$ represents the relationship between $x$ and $y$.

**Choice A** is incorrect. The point $(0,40)$ is not on the graph of this equation, since $40=8(0)+12$, or $40=12$, is not true. **Choice C** is incorrect. The point $(0,40)$ is not on the graph of this equation, since $40=12(0)+8$, or $40=8$, is not true. **Choice D** is incorrect. The point $(0,40)$ is not on the graph of this equation, since $12(0)+8(40)=480$, or $320=480$, is not true.

**Figure 5. Example of GPT-4o Correct Answer on SAT Math: Question (Top), GPT-4o Response (Bottom Left), and Correct Answer (Bottom Right)**

The perimeter of an equilateral triangle is 624 centimeters. The height of this triangle is $k\sqrt{3}$ centimeters, where $k$ is a constant. What is the value of $k$?

**Solution:**

First, find the side length of the equilateral triangle:
Perimeter $= 3 \times$ side length
$624 = 3 \times$ side length
side length $= \frac{624}{3} = 208$ cm

For an equilateral triangle with side length $s$, the height $h$ is given by:
$h = \frac{\sqrt{3}}{2} \times s$
$k\sqrt{3} = \frac{\sqrt{3}}{2} \times 208$
$k = \frac{208}{2} = 104$

**Answer:** 104

---

The correct answer is 104. An equilateral triangle is a triangle in which all three sides have the same length and all three angles have a measure of 60°. The height of the triangle, $k\sqrt{3}$, is the length of the altitude from one vertex. The altitude divides the equilateral triangle into two congruent 30-60-90 right triangles, where the altitude is the side across from the 60° angle in each 30-60-90 right triangle. Since the altitude has a length of $k\sqrt{3}$, it follows from the properties of 30-60-90 right triangles that the side across from each 30° angle has a length of $k$ and each hypotenuse has a length of $2k$. In this case, the hypotenuse of each 30-60-90 right triangle is a side of the equilateral triangle; therefore, each side length of the equilateral triangle is $2k$. The perimeter of a triangle is the sum of the lengths of each side. It's given that the perimeter of the equilateral triangle is 624; therefore, $2k+2k+2k=624$, or $6k=624$. Dividing both sides of this equation by 6 yields $k=104$.

## Performance on MBE

The Multistate Bar Examination (MBE) is a standardized test that assesses the ability of prospective lawyers to apply fundamental legal principles and reasoning to analyze given fact patterns (National Conference of Bar Examiners, n.d.). Developed and administered by the National Conference of Bar Examiners (NCBE), the MBE is a critical component of the bar examination in most U.S. jurisdictions (Griggs, 2019). The MBE includes 200 multiple-choice questions that cover a wide range of legal topics, including constitutional law, contracts, evidence, real property, and torts. The test evaluates the examinee's capacity to think like a lawyer and apply legal knowledge in a practical, problem-solving context.

The dataset for evaluating GPT-4o's performance includes sample test questions from the MBE sample booklet, updated in 2023[4]. These questions represent the types and formats of questions that examinees will encounter on the actual MBE, providing a comprehensive overview of the subjects tested and the skills required. In this test, GPT-4o correctly answered 15 out of 20 questions, leading to 75% accuracy. The comparison with previous models is presented in Table 5, based on the results reported by Katz et al. (2024).

---

[4] https://www.ncbex.org/sites/default/files/2023-05/MBE_Sample_Test_Questions_New_2023%20.pdf

| Model | Accuracy |
|---|---|
| GPT-3.5 | 45.10% |
| GPT-4 | 75.70% |
| GPT-4o | 75.00% |

Table 5. Performance Comparison of GPT models on MBE

The evaluation results indicate that GPT-4o performs comparably to GPT-4 on the MBE, with a minor difference in accuracy. However, compared to GPT-3.5, which achieved an accuracy of 45.10%, GPT-4o demonstrates a significant improvement. Therefore, law students and bar examinees can benefit from using GPT-4o as an interactive learning tool that provides immediate feedback and explanations, helping them to understand complex legal principles and improve their problem-solving skills.

## 2.2 Reasoning

Human intellect is remarkably characterized by reasoning, which is described as an activity of methodically and logically thinking about a subject (Huang & Chang, 2023). Reasoning enables humans to come to conclusions or make decisions by using previous experiences and data gathered, thus extending one's knowledge of the world and releasing the possibility for innovation and development. In recent times, AI has made significant advancements in narrowing the gap between human and machine intellect through the use of Natural Language Processing (NLP) and LLMs, which have established remarkable reasoning abilities.

**Figure 6. Types of Reasoning**

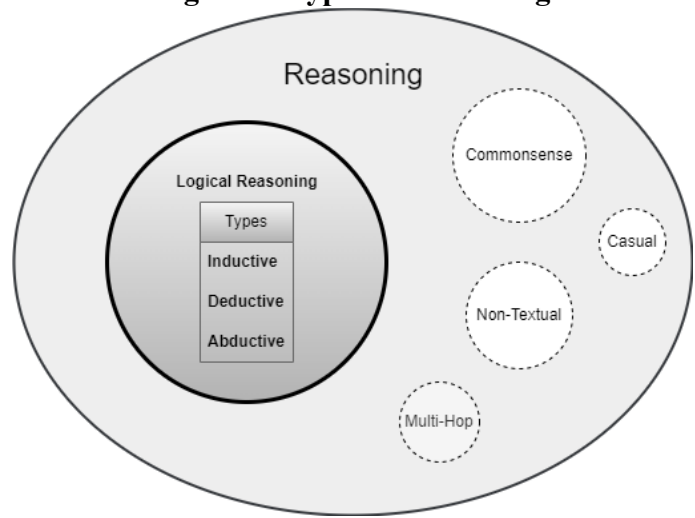

In this section, the authors assess the reasoning capacity of the most recent GPT-4o model by manual technical evaluation through a sequence of question-answering tasks. The model will answer a range of logical reasoning tasks in different types, including deductive, inductive, and abductive reasoning, as shown in Figure 6. Starting with a broad principle or assumption and applying it to produce predictions or draw conclusions, deductive reasoning takes a top-down method (Johnson-Laird, 2010). By contrast, inductive reasoning uses a bottom-up methodology to deduce broad principles or conclusions from observations or data (Hayes et al., 2010). In abductive reasoning, theories or explanations are developed from little, ambiguous, or incomplete data (Walton, 2014). With all these assessments on the model, this article stands to obtain an understanding of the reasoning capacity of the GPT-4o model in various settings.

**Figure 7. Logical Reasoning Categories and Datasets**

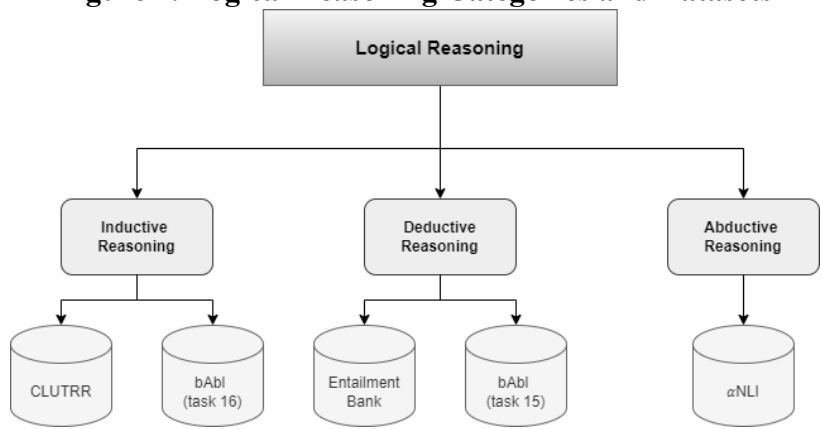

In this subsection, we assess the performance of GPT-4o on five datasets that include all the aforementioned types of reasoning, as illustrated in Figure 7. To evaluate the deductive reasoning ability, two datasets were utilized namely EntailmentBank (Dalvi et al., 2021) and bAbI (task 15) (Weston et al., 2015). Similarly, to assess the capability of inductive reasoning, we employed two datasets CLUTRR (Sinha et al., 2019) and bAbI (task 16) (Weston et al., 2015). For abductive reasoning, we use the αNLI dataset (Bhagavatula et al., 2019). Adhering to the methods of López Espejel et al. (2023), our evaluation encompassed the same set of 30 randomly chosen samples from each evaluation dataset. The observations are selected from 10 samples from each of the training-easy, train-medium, and train-hard sets of the $\alpha$NLI dataset. Concurrently, a total of 30 samples are drawn from the test set for the bAbI (task 15), bAbI (task 16), CLUTRR, and EntailmentBank datasets. In accordance with López Espejel et al. (2023), we utilized the identical set of proven and effective prompts that were implemented in their assessment to evaluate the capabilities of the model.

| Category/ Model | **Deductive Reasoning** | | **Inductive Reasoning** | | **Abductive Reasoning** |
|---|---|---|---|---|---|
| | Entailment Bank | bAbI (task 15) | CLUTRR | bAbI (task 15) | αNLI |
| GPT 3.5 | 25/30 | 26/30 | 2/30 | 14/30 | 19/30 |
| GPT 4 | 27/30 | 30/30 | 11/30 | 28/30 | 25/30 |
| GPT 4o | 29/30 | 30/30 | 17/30 | 30/30 | 27/30 |

Table 6. Performance of GPT Models on Logical Reasoning Tasks

The evaluation results showcase the remarkable reasoning abilities of GPT-4o in all three domains, as indicated in Table 6. GPT-4o demonstrated exceptional performance in deductive reasoning by achieving nearly flawless scores on both bAbI (task 15) and EntailmentBank. It outperformed ChatGPT-3.5 and performed at the same level as ChatGPT-4 (López Espejel et al., 2023). GPT-4o achieved flawless results in the inductive reasoning tasks, scoring perfectly on bAbI (task 16) and achieving a score of 17 out of 30 on CLUTRR. It outperformed both ChatGPT-3.5 and ChatGPT-4. GPT-4o achieved a score of 27 out of 30 on αNLI, surpassing the performance of its previous versions in abductive reasoning. The results underscore GPT-4o's superior reasoning capabilities compared to its predecessors. The model's proficiency in deductive reasoning showcases its ability to derive valid conclusions from premises. Its success in inductive reasoning demonstrates the capacity to generalize from specific facts, while its performance in abductive reasoning highlights the ability to generate credible hypotheses with limited knowledge.

Even with the remarkable reasoning powers of GPT-4o, this assessment points to a few drawbacks that need more rigorous investigation. One case was when the model gave different answers to the same topic in various chat sessions while evaluating the bAbI (task 16) dataset for inductive reasoning. Additionally, the model sometimes requested the end-user to choose between different answers. This implies that in some situations, GPT-4o could have trouble with ambiguity, resulting in varying responses. Furthermore, the model sometimes gave different responses when the same subject was posed repeatedly in the same chat session. In determining accuracy, only the first response was considered to maintain synchrony, although this inconsistency raises questions about the model's efficiency and dependability in certain areas. The model's sensitivity to question-wording, the information presentation sequence, or the existence of unclear or contradictory information in the input may be the causes of these issues. To overcome these problems and enhance its capacity to manage ambiguity and resolve contradictions, future studies should concentrate on creating more reliable and consistent reasoning mechanisms and optimizing prompts for LLMs.

With the notable performance of GPT-4o, there might be more advancements for many AI applications with GPT-4o's improved reasoning skills. High-accuracy complex reasoning tasks performed by it can lead to advancements in information retrieval, decision support tools, and question-answering systems. Still, further study is required to determine how well the model performs on a larger variety of reasoning problems and how well it can manage more intricate and domain-specific reasoning situations.

## 2.3 Language Translation

Language translation has become an increasingly important task in our globalized world, facilitating communication and understanding across diverse linguistic backgrounds. With the advent of LLMs like GPT 3.5, GPT 4, Llama, Gemini and now GPT-4o, the potential for accurate and efficient machine translation has grown significantly (Khoshafah, 2023). These models, which were trained on massive volumes of multilingual data, can produce translations accurately while capturing the subtle meanings and complexities of different languages. Therefore, in this section, we aim to evaluate the translation proficiency of GPT-4o in six of the most widely spoken languages: Spanish, Arabic, Hindi, French, Portuguese, and Russian.

The choice of these six languages is not arbitrary; they represent a diverse set of linguistic structures and cultural contexts, making them ideal for a comprehensive evaluation of translation capabilities. Spanish is commonly used across Europe and the Americas and is characterized by its straightforward structure and rich vocabulary. Arabic, known for its intricate script and complex word forms, poses distinct challenges for translation technology. Hindi, widely spoken in India, mixes local and foreign words, requiring careful handling to achieve accurate translation. French, spoken in many parts of the world, helps test the model's ability to handle grammatical rules and nuances. Portuguese, similar to Spanish but distinct in several key aspects, allows for an assessment of the model's precision in closely related languages. Lastly, Russian, with its Cyrillic script and case system, provides a test for the model's ability to manage non-Latin scripts and complex grammatical structures.

By focusing on these languages, this study aims to provide a robust and diverse evaluation of GPT-4o's translation performance. Given the widespread use and significant number of native speakers of these languages, improvements in translation accuracy can have a substantial impact on global communication and information dissemination. Hence in this section, we seek to verify GPT-4o's ability to translate across these six languages, providing insights into its potential for breaking down language barriers and facilitating communication among people from different linguistic backgrounds.

**Data**

The datasets for Spanish, Arabic, French, Portuguese, and Russian were sourced from the OPUS dataset, a well-known collection of texts used for training and evaluating machine translation models (Tiedemann, 2012), and the Hindi dataset was obtained from the IIT Bombay English-Hindi Parallel Corpus, created by the Center for Indian Language Technology (CFILT) at IIT Bombay (Kunchukuttan et al., 2018).

For this analysis, 500 data points were randomly sampled from each dataset. The selection of 500 data points is a good balance between feasibility and the need for sufficient data diversity. This sample size is large enough to encompass a wide variety of sentence structures, vocabulary, and translation challenges present in each language, ensuring that the evaluation is comprehensive and representative. Random sampling was employed to mitigate selection bias and to ensure that the sampled data points provide an unbiased representation of the overall dataset. By using random sampling, this approach captures the natural variability and complexity of language, which is essential for a robust assessment of the GPT-4o model's translation performance across different linguistic contexts.

**Evaluation Method**

To measure how similar two sentences are in terms of their meaning, an advanced NLP, specifically focusing on sentence embeddings generated by a model called BERT (Bidirectional Encoder Representations from Transformers) (Devlin et al., 2019) and a similarity measure called cosine similarity has been used. BERT is a powerful model that has greatly improved how well computers understand language. For our research, we use a pre-trained model from the sentence-transformers library called paraphrase-MiniLM-L6-v2. This model is specially tuned to understand the meanings and similarities between sentences. It works by turning each sentence into a vector, which is a list of numbers. These vectors or embeddings, encapsulate the semantic information of the sentences in a way that allows for meaningful comparison between the actual translations and the translations generated by GPT 4o.

To find out how similar two sentences are, we compare their vectors using cosine similarity. Cosine similarity measures the angle between two vectors or embeddings. If the vectors point in the same direction, the sentences are very similar. If they point in completely different directions, the sentences are very different. The values are between -1 and 1, where:

- 1 indicates that the vectors are identical.
- 0 indicates that the vectors are orthogonal (i.e., no similarity).
- -1 indicates that the vectors are opposed.

By calculating the cosine similarity between the embeddings of two sentences, we can effectively measure their semantic similarity. The formula for cosine similarity is:

Cosine Similarity = $\mathbf{A} \cdot \mathbf{B} / \|\mathbf{A}\| \|\mathbf{B}\|$

where **A** and **B** are the embeddings of the two sentences.

**Results**

This study sought to evaluate the capabilities of GPT-4o in translating passages across six major languages: Spanish, Arabic, Hindi, French, Portuguese, and Russian. The results reveal a generally high level of translation accuracy, particularly in Spanish and Portuguese, which scored 88% and 86% respectively. However, there were notable variations among the languages. Arabic and French, with scores of 78% and 75%, respectively, presented more challenges for the model due to their complex linguistic structures and nuances. Hindi and Russian scored 82% and 80%, demonstrating the model's competence but also highlighting areas for improvement. The results are summarized in Table 7.

| Language | Translation Accuracy (%) |
|---|---|
| Spanish | 88 |
| Arabic | 78 |
| Hindi | 82 |
| French | 75 |
| Portuguese | 86 |
| Russian | 80 |

Table 7. GPT-4o Translation Accuracy Across Languages

The findings suggest that the line between human and machine translation is becoming increasingly narrow. GPT-4o's performance, though not specifically optimized for translation, approaches the quality of dedicated translation systems. This is particularly noteworthy given the diverse linguistic and structural characteristics of the evaluated languages. While the exact nature of the source translations in the datasets (whether human or machine-translated) is not confirmed, the high similarity scores indicate that GPT-4o is capable of producing translations with a quality that is comparable to the existing translations. However, several limitations must be considered. The random sampling of 500 data points from each dataset may not fully capture the linguistic diversity and complexity of each language. Different samples could yield varying results, suggesting that a larger and more representative dataset might provide a more accurate assessment. Additionally, the reliance on BERT-based embeddings and cosine similarity may not fully encapsulate the nuances of translation quality, particularly in capturing cultural and contextual subtleties. Expanding the dataset size and including more language pairs could yield more comprehensive insights. This research serves as a proof-of-concept for larger-scale studies that could further investigate the capabilities of AI in translation. Future research should focus on incorporating more extensive data, diverse language combinations, and advanced fine-tuning techniques.

## 3. Vision Capacity of GPT-4o

Vision capacity is foundational to developing intelligent models capable of understanding, interpreting, and interacting with visual content. This capacity encompasses a range of skills that enable models to process and produce coherent and contextually appropriate responses to visual inputs. The objective is to comprehensively assess the vision performance of GPT-4o by testing it on various image-based tasks. Each of these tasks is significant for evaluating different aspects of the model's visual capabilities.

For each task, a dataset of approximately 100 representative images was curated. The model was provided with an image along with a text prompt specifying the desired output format. The prompts were designed to probe the model's ability to identify, classify, describe, and analyze visual content without additional context. For select tasks, we further investigated the model's few-shot learning capabilities by providing a small number of labeled examples before the query image.

Model outputs were compared against ground truth labels to compute standard performance metrics such as accuracy. Qualitative analysis was also conducted on a subset of responses to identify common failure modes and strengths. The results across different tasks provide insights into GPT-4o's current visual understanding capabilities, areas for improvement, and potential as a foundation model for vision tasks. Subsequent sections discuss the specifics of each task, dataset, and findings, offering a comprehensive evaluation of GPT-4o's visual reasoning skills.

### 3.1 Fruits Classification

Fruit image classification is crucial for applications in agriculture, supply chain management, and food industry automation. Accurate identification of fruit types can enhance inventory tracking, quality control, and efficient sorting processes (Cubero et al., 2011). The fruit images dataset[5] consists of approximately 400 images spanning 10 different fruit classes such as banana, jackfruit, and mango. Each fruit class has 40 labeled images, with the dataset split into 320 training images and 80 test images. The images were collected from various sources, such as Google Images and stock image websites, and were labeled by the dataset creators. For this evaluation, the model was provided with an image along with a prompt to identify the fruit class from the list of 10 classes in a specified format. Model predictions were compared against ground truth labels to assess performance.

The results indicate that GPT-4o performed exceptionally well on this task. The model achieved an average **precision of 0.98**, an average **recall of 0.98**, and an average **F1-score of 0.98**. These metrics suggest that GPT-4o is highly capable of accurately identifying and classifying different fruit images. Table 8 summarizes the performance for each class.

---

[5] https://www.kaggle.com/datasets/afsananadia/fruits-images-dataset-object-detection

| Class | Precision | Recall | F1-Score |
|---|---|---|---|
| Banana | 0.91 | 1.00 | 0.95 |
| Papaya | 1.00 | 1.00 | 1.00 |
| Apple | 1.00 | 1.00 | 1.00 |
| Litchi | 1.00 | 1.00 | 1.00 |
| Jackfruit | 1.00 | 0.90 | 0.95 |
| Hog Plum | 1.00 | 1.00 | 1.00 |
| Grapes | 1.00 | 1.00 | 1.00 |
| Guava | 1.00 | 1.00 | 1.00 |
| Mango | 0.91 | 1.00 | 0.95 |
| Orange | 1.00 | 0.90 | 0.95 |

Table 8. GPT-4o Performance on Fruit Classification

The model demonstrated strong performance in classifying the 10 different fruit classes, achieving high precision, recall, and F1-scores across most classes. Several classes, including Papaya, Apple, Litchi, Hog Plum, Grapes, and Guava, obtained perfect scores of 1.0 for precision, recall, and F1-score. The Banana and Mango classes had slightly lower but still impressive precision scores (0.91), with a perfect recall of 1.0. Figure 8 presents the confusion matrix and metric visualization for this dataset.

**Figure 8. Confusion Matrix (left) and Performance Comparison (right) for Fruits Classification**

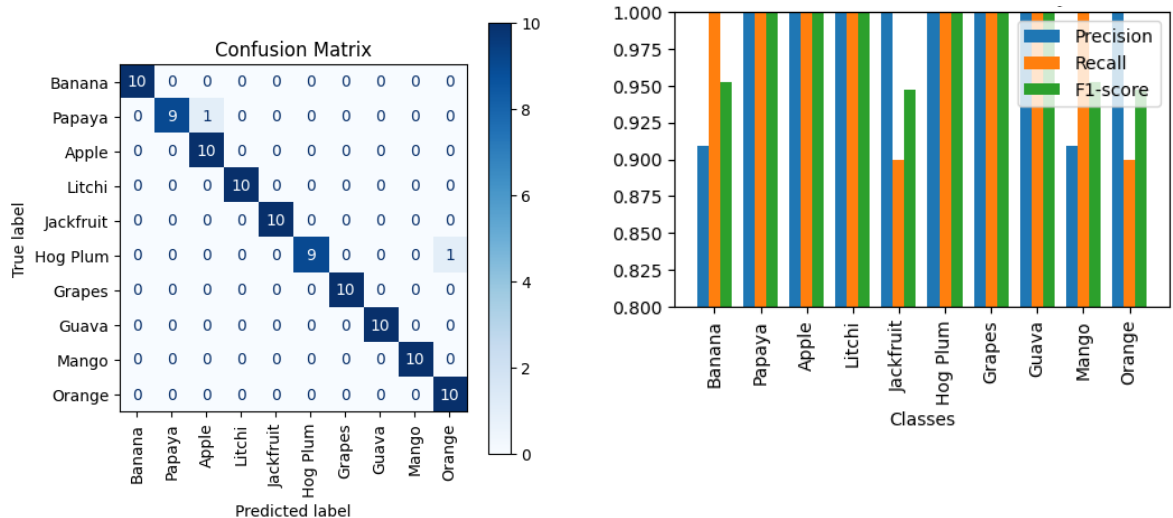

### 3.2 Driver Drowsiness Detection

Detecting driver drowsiness is critical for enhancing road safety, as timely identification of fatigue can prevent accidents and save lives. The drowsy detection dataset consists of images extracted from videos capturing drivers in three distinct states: natural, fatigued, and drowsy

(Jebraeily et al., 2024). The dataset was curated by gathering relevant videos, converting them into image frames, and applying facial detection algorithms to isolate key facial regions like eyes, mouth, and cheeks, which are indicative of drowsiness. The extracted images were converted to grayscale, resized to 48x48 pixels, and accurately labeled based on the driver's state. The dataset comprises two classes: drowsy and natural, with a total of 100 labeled images sampled evenly from each class. For this evaluation, GPT-4o was provided with an image along with a prompt to classify it into one of the two classes in a specified JSON format. The model's predictions were compared against the ground truth labels to assess its performance in detecting driver drowsiness from facial features.

In this task, the model achieved an average **precision of 0.80**, an average **recall of 0.80**, and an average **F1-score of 0.80.** Table 9 summarizes the performance for each class.

| Class   | Precision | Recall | F1-Score |
|---------|-----------|--------|----------|
| Drowsy  | 0.8       | 0.8    | 0.8      |
| Natural | 0.8       | 0.8    | 0.8      |

Table 9. GPT-4o Performance on Drowsiness Detection

The results indicate that GPT-4o, without fine-tuning, achieves an impressive precision, recall, and F1-score of 0.8. While lower than that of specialized deep learning models like VGG, ResNet, and CNN (Jebraeily et al., 2024), the performance is impressive given GPT-4o's lack of training on this specific dataset. The notable performance despite no domain-specific training underscores its robustness and adaptability, implying that GPT-4o could be valuable in scenarios where rapid deployment and flexibility across different tasks are crucial. Figure 9 presents the confusion matrix and metric visualization for this task.

**Figure 9. Confusion Matrix (left) and Performance Comparison (right) for Drowsiness Detection**

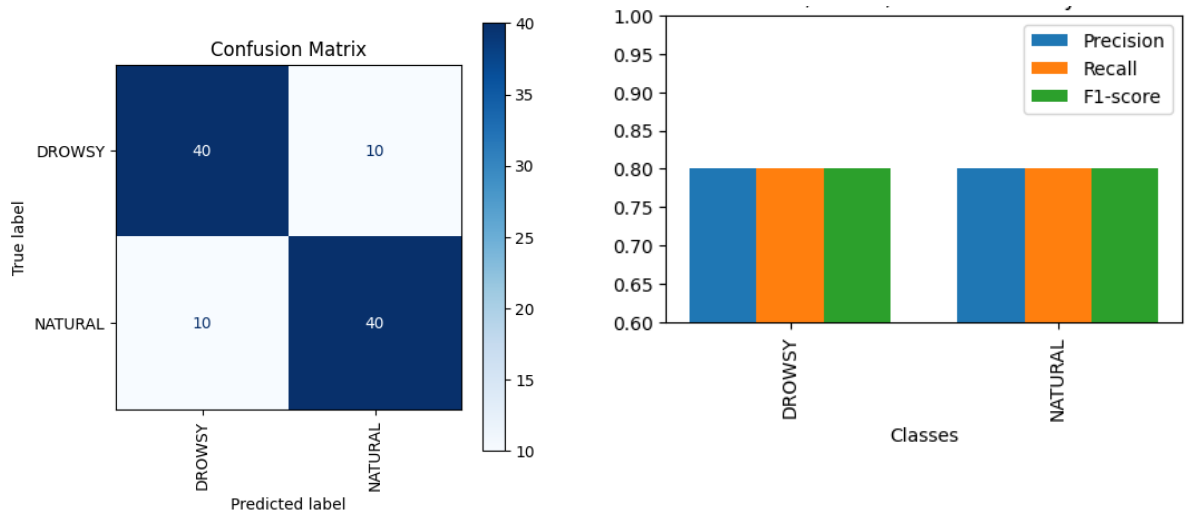

## 3.3 Crop Disease Classification

Accurate identification of crop diseases is essential for ensuring agricultural productivity and preventing significant crop losses. The crop disease classification dataset is a comprehensive collection of images aimed at evaluating GPT-4o's capabilities in identifying crop diseases[6]. The dataset encompasses 20 distinct classes of common crop diseases, including blight, cedar apple rust, crown gall, and clubroot. For this evaluation, 100 images were randomly sampled from the dataset, with each class represented by approximately five images. GPT-4o was provided with these images along with a prompt to classify the crop disease depicted in each image. The model's predictions were compared against the ground truth labels to assess its performance in accurately identifying and distinguishing various crop diseases based solely on visual information.

The model achieved an average **precision of 0.77**, an average **recall of 0.71**, and an average **F1-score of 0.68** in this task. Table 10 summarizes the performance for each class.

| Class | Precision | Recall | F1-score |
| --- | --- | --- | --- |
| Anthracnose | 0.60 | 0.60 | 0.60 |
| Apple Scab | 1.00 | 0.80 | 0.89 |
| Black Spot | 0.67 | 1.00 | 0.80 |
| Blight | 0.38 | 0.75 | 0.50 |
| Blossom End Rot | 1.00 | 1.08 | 1.00 |
| Botrytis | 1.00 | 0.20 | 0.33 |
| Brown Rot | 1.00 | 0.20 | 0.33 |
| Canker | 0.25 | 0.25 | 0.25 |
| Cedar Apple Rust | 0.83 | 1.00 | 0.91 |
| Clubroot | 1.00 | 1.00 | 1.00 |
| Crown Gall | 1.00 | 1.00 | 1.00 |
| Downy Mildew | 1.00 | 0.20 | 0.33 |
| Fire Blight | 0.80 | 0.80 | 0.80 |
| Fusarium | 1.00 | 0.60 | 0.75 |
| Gray Mold | 0.43 | 0.75 | 0.55 |
| Leaf Spots | 0.40 | 0.80 | 0.53 |
| Mosaic Virus | 0.67 | 0.80 | 0.72 |
| Nematodes | 1.00 | 1.00 | 1.00 |
| Powdery Mildew | 0.71 | 1.00 | 0.83 |
| Verticillium | 0.67 | 0.40 | 0.50 |

Table 10. GPT-4o Performance on Crop Disease Detection

Given the large number of classes (20), this highlights GPT-4o's potential for accurate crop disease classification and adaptability, despite no prior training on this dataset. The limitations in specific classes like Botrytis, Brown rot, and Canker can be attributed to the need for specialized training in certain classes. The confusion matrix and metric visualization for this dataset are presented in Figure 10.

---

[6] https://www.kaggle.com/datasets/sadikaljarif/crop-disease

**Figure 10. Confusion Matrix (top) and performance comparison (bottom) for Crop Disease Classification**

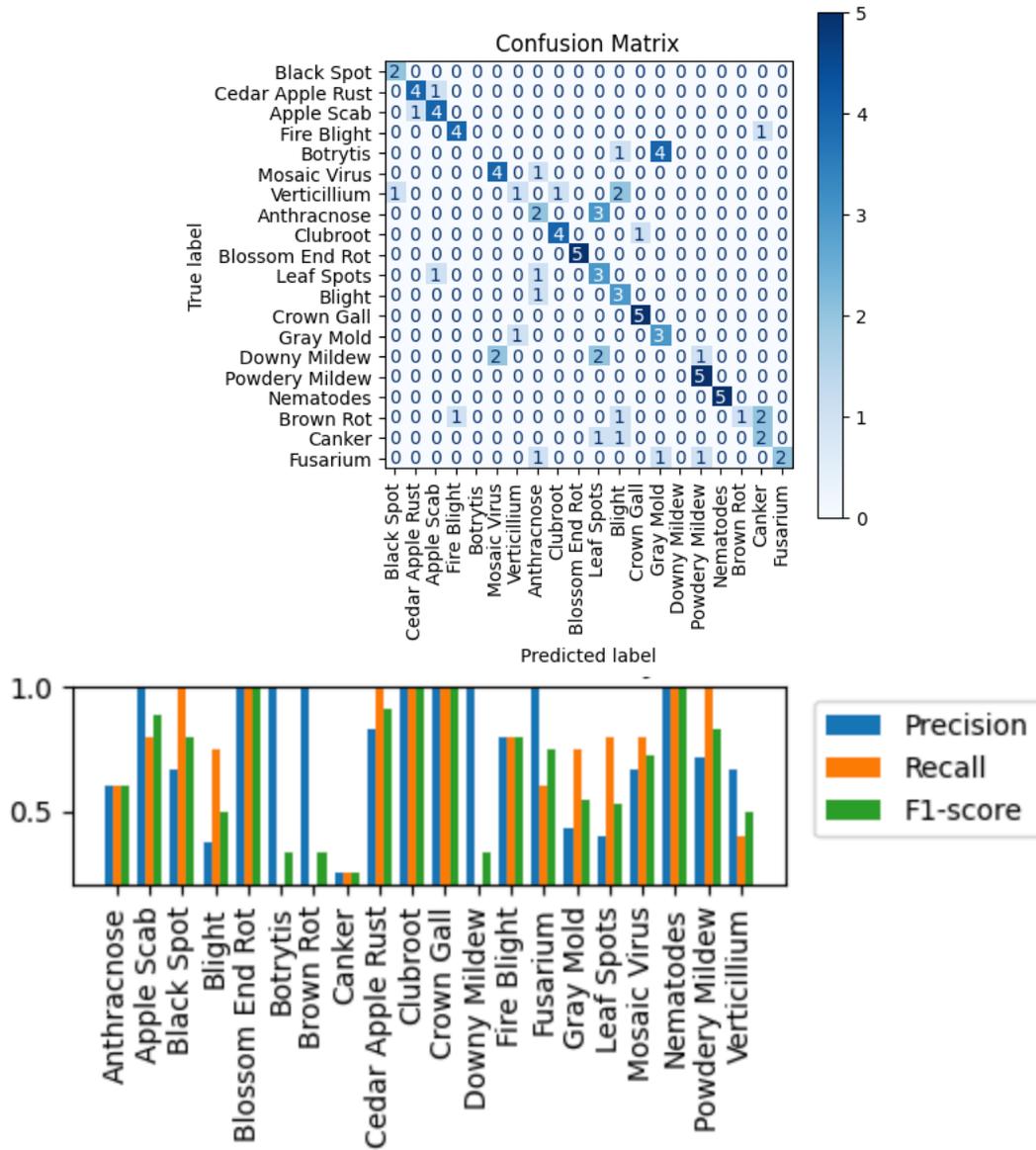

### 3.4 Glaucoma Detection

Early detection of glaucoma is critical for preventing vision loss and ensuring timely treatment. The glaucoma detection dataset used for this evaluation consisted of retinal fundus images from the ACRIMA database[7]. A subset of 100 images was sampled, evenly split between glaucomatous and normal cases. These images were collected at FISABIO Oftalmología Médica in Valencia, Spain, and were annotated by experienced glaucoma experts. GPT-4o was tasked with classifying each image into either glaucoma or normal based solely on the visual information provided. The model's predictions were compared against the expert-annotated ground truth labels to assess its performance in detecting glaucoma from retinal fundus imagery.

---

[7] https://www.kaggle.com/datasets/chetanpediredla/glaucoma-dataset

As shown in Table 11, GPT-4o achieved an average **precision of 0.65**, an average **recall of 0.62**, and an average **F1-score of 0.59**. For the glaucoma class, the model demonstrated a precision of 0.58, a recall of 0.86, and an F1-score of 0.69. In contrast, the normal class had a higher precision of 0.73 but a significantly lower recall of 0.38, resulting in an F1-score of 0.50.

| Class | Precision | Recall | F1-Score |
|---|---|---|---|
| Glaucoma | 0.58 | 0.86 | 0.69 |
| Normal | 0.73 | 0.38 | 0.50 |

Table 11. GPT-4o Performance on Glaucoma Detection

The confusion matrix in Figure 11 reveals that the model correctly identified 42 out of 49 glaucoma cases but struggled more with normal cases, correctly classifying only 19 out of 50. The plot shows the model's relatively balanced precision and recall for the glaucoma class but highlights a pronounced discrepancy for the normal class, with precision substantially higher than recall. GPT-4o is effective at identifying glaucomatous images but has difficulty in correctly classifying normal cases.

**Figure 11. Confusion Matrix (left) and Performance Comparison (right) for Glaucoma Detection**

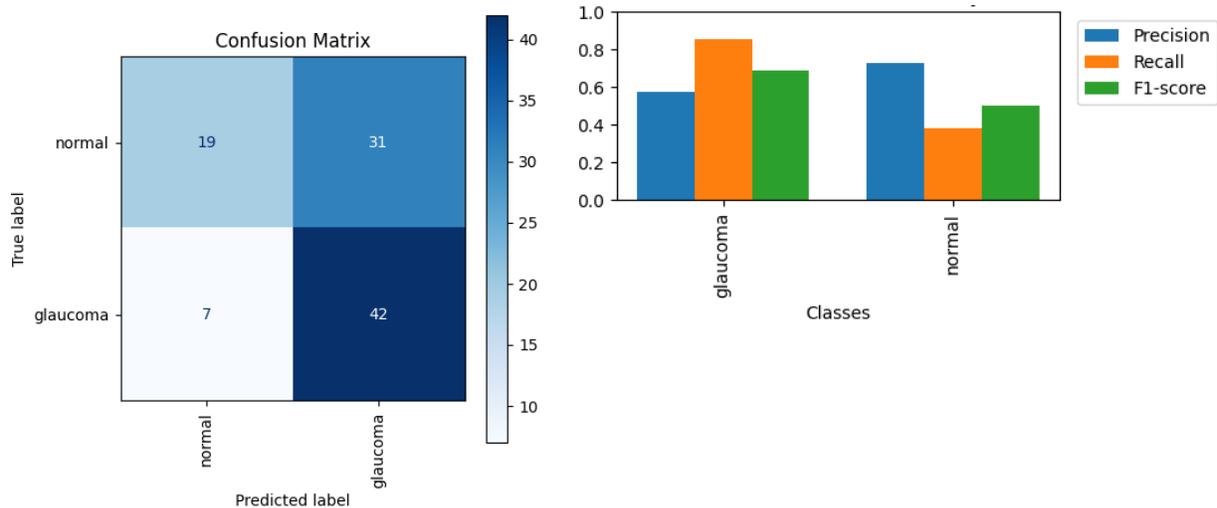

Few-shot learning allows models to make accurate predictions with only a small number of training examples. This approach is particularly beneficial in scenarios where data is scarce. Figure 12 illustrates the F1 scores for both classes across different numbers of shots, indicating how the model's performance evolves with the number of examples provided during training. The glaucoma class maintains a relatively high F1 score across all shot levels, showing slight improvement with additional examples. This consistency suggests that GPT-4o effectively learns to identify glaucomatous features even with a limited number of examples. In contrast, the normal class exhibits significant improvement in F1 score from zero shots to one shot but then plateaus. This indicates that while the initial provision of examples significantly enhances the model's ability to recognize normal cases, further increases in the number of examples yield diminishing returns.

**Figure 12. Performance Evolution Against Number of Shots for Glaucoma Detection**

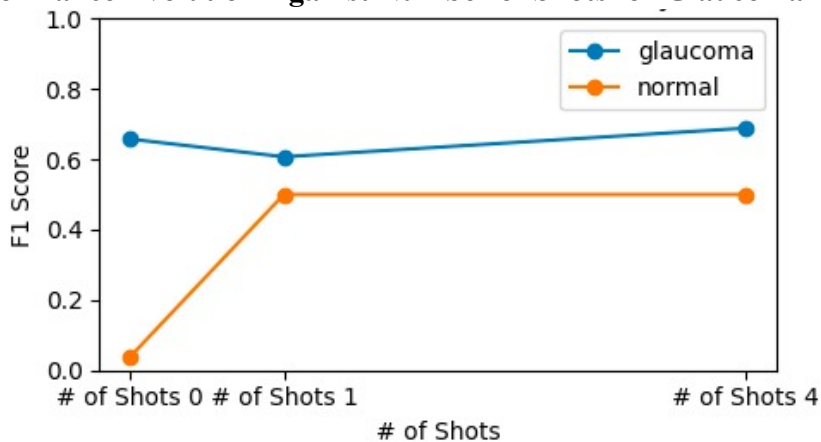

## 3.5 Cancer, Tumor, and Aneurysm Detection

Accurate detection and classification of brain conditions such as cancer, tumors, and aneurysms are crucial for timely diagnosis and treatment. The computed tomography (CT) brain scan dataset contains CT images of the brain aimed at detecting and classifying various conditions such as cancer, tumors, and aneurysms[8]. For this evaluation, a subset of 100 CT scan images was sampled from the dataset. GPT-4o was tasked with analyzing these images and classifying them into one of three categories: cancer, tumor, or aneurysm. The model's predictions were compared against the ground truth labels to assess its performance in identifying these medical conditions from CT brain imagery.

GPT-4o achieved an average **precision of 0.21**, an average **recall of 0.32**, and an average **F1-score of 0.26.** Table 12 summarizes the performance for each class.

| Class | Precision | Recall | F1-Score |
|---|---|---|---|
| Aneurysm | 0.3725 | 0.5758 | 0.4524 |
| Cancer | 0.0000 | 0.0000 | 0.0000 |
| Tumor | 0.2708 | 0.3939 | 0.3210 |

Table 12. GPT-4o Performance on Cancer, Tumor, and Aneurysm Detection

The confusion matrix in Figure 13 reveals that the model completely failed to predict the 'cancer' class, potentially due to a lack of representative training data or inherent similarities with other classes. Additionally, it struggled to distinguish between 'aneurysm' and 'tumor' classes, with significant misclassifications in both directions, suggesting a need for further fine-tuning or incorporation of additional relevant features.

---

[8] https://www.kaggle.com/datasets/trainingdatapro/computed-tomography-ct-of-the-brain

**Figure 13. Confusion Matrix for Cancer, Tumor, and Aneurysm Detection Task**

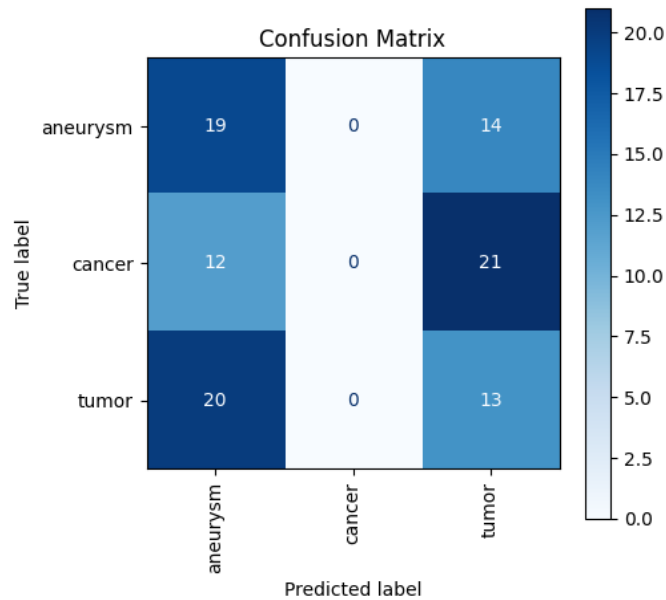

### 3.6 Image Captioning

The flickr8k captions dataset is a compact collection designed for image captioning tasks[9]. It consists of 8,000 images sourced from Flickr, each accompanied by multiple human-annotated captions describing the visual content. For this evaluation, a subset of 100 images from the dataset was used. GPT-4o was tasked with inferring the context and generating natural language descriptions for these 100 images. The model's generated captions were compared against the ground truth human captions using the BLEU score, a metric that measures the similarity between machine and human-generated texts. This dataset helps assess the ability to comprehend visual scenes and translate them into accurate and coherent textual descriptions. The BLEU scores obtained are summarized in Table 13.

| Metric | Score |
|--------|-------|
| BLEU-1 | 0.193 |
| BLEU-2 | 0.095 |
| BLEU-3 | 0.058 |
| BLEU-4 | 0.031 |

Table 13. GPT-4o Performance on Image Captioning

With a BLEU-1 score of 0.193, the model demonstrates a moderate ability to capture the essence of the captions with a reasonable degree of similarity in individual words. However, as the n-gram length increases, the scores decline significantly (BLEU-2: 0.095, BLEU-3: 0.058, BLEU-4: 0.031), indicating that the model struggles with maintaining coherence and context in longer sequences. This highlights the challenges GPT-4o faces in generating more complex and accurate descriptions. The results show that GPT-4o has a foundational understanding of visual scenes, but there is room for improvement in generating detailed and contextually rich captions.

---

[9] https://www.kaggle.com/datasets/aladdinpersson/flickr8kimagescaptions

## 4. Speech Capacity of GPT-4o

Speech capacity evaluates the ability of intelligent models to understand, interpret, and interact with auditory content. This encompasses a range of skills that enable models to process and produce coherent and contextually appropriate responses to audio inputs. The objective is to assess the audio performance of GPT-4o by testing it on various audio-based tasks. Each of these tasks is significant for evaluating different aspects of the model's auditory capabilities.

### 4.1 Emotion Detection

Emotion detection is a critical aspect of understanding human communication, as the same speech can convey different meanings depending on the emotional tone in which it is expressed (Shahriar et al., 2023). Recognizing emotions in speech is essential for applications ranging from customer service to mental health monitoring. For this evaluation, we used the Arabic natural audio dataset (ANAD) from Kaggle, designed to detect discrete emotions in Arabic speech[10]. The ANAD consists of 1,384 audio recordings, each labeled with one of three emotions: happy, angry, or surprised. These recordings were sourced from live Arabic talk shows, where each video was labeled by 18 listeners to determine the perceived emotion. To evaluate the emotion detection capabilities of GPT-4o, we randomly sampled 100 audio files from the ANAD dataset. Each audio file was fed to the model along with a prompt to predict the emotion class. The model's predictions were then compared against the ground truth labels to assess its performance.

The results of the emotion detection task, as illustrated in Figure 14, reveal that GPT-4o demonstrates variable performance across different emotion classes. The confusion matrix shows that the model performs best for the "surprised" class, correctly predicting 21 instances, but it frequently misclassifies "happy" as "surprised" (19 times). The "angry" class has the lowest true positive rate with only two correct predictions, often being mistaken for "happy" or "surprised." The model has the highest recall for the "surprised" class, indicating it correctly identifies "surprised" emotions more frequently than others. The precision for "angry" is reasonably high, but the recall is very low, meaning that while it predicts "angry" correctly when it does so, it rarely predicts "angry" overall. The "happy" class has moderate precision and recall, suggesting a balanced but moderate performance in predicting this class.

---

[10] https://www.kaggle.com/datasets/suso172/arabic-natural-audio-dataset/data

# Figure 14. Confusion Matrix (top) and Performance Comparison (bottom) for Audio Emotion Detection

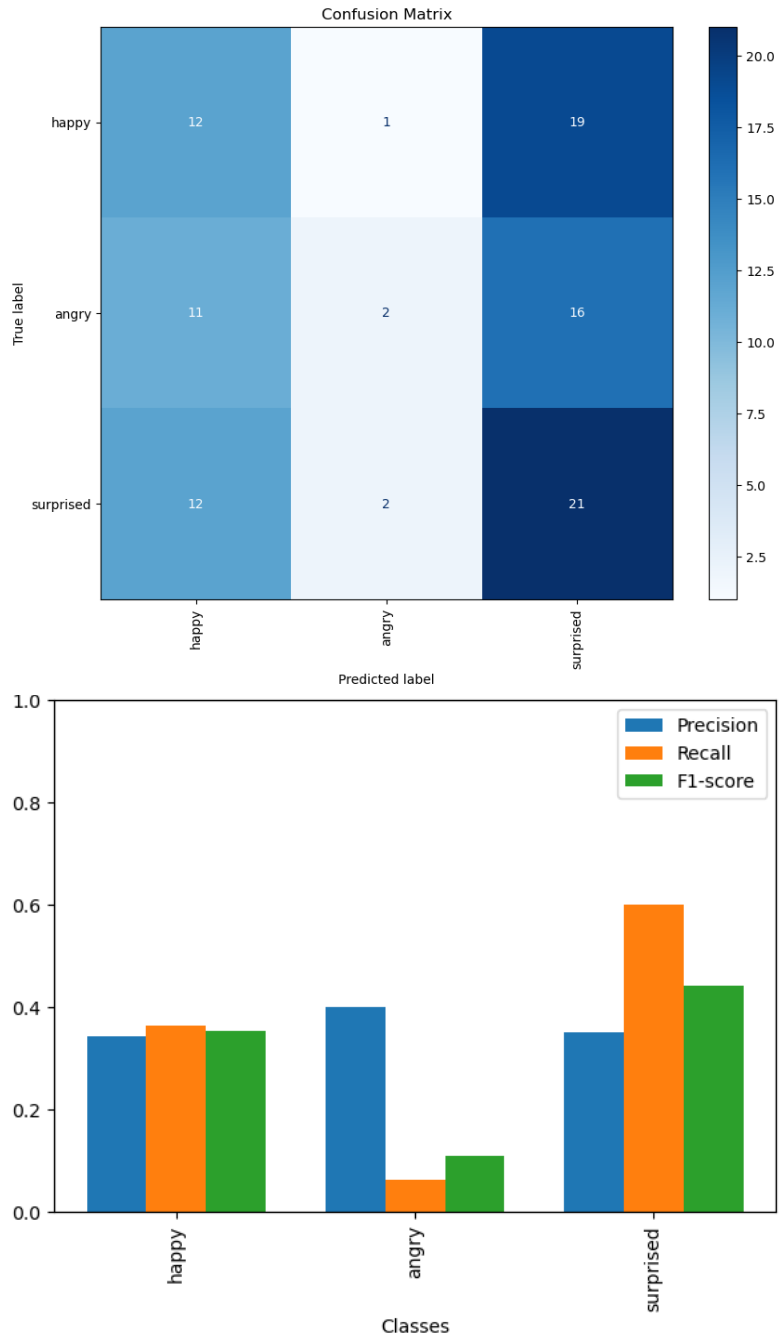

## 4.2 Accent Detection

Accents play a crucial role in speech recognition, affecting the accuracy and efficiency of automatic speech recognition (ASR) systems. Understanding and detecting accents is essential for developing robust ASR systems that can handle diverse linguistic backgrounds (Graham & Roll, 2024). For this evaluation, we utilized the AccentDB dataset, a comprehensive collection of non-native English accents designed to assist neural speech recognition tasks (Ahamad et al., 2020).

The AccentDB dataset includes samples from various Indian-English accents and native English accents, providing a diverse range of phonetic and prosodic variations. It contains speech recordings from speakers with distinct linguistic backgrounds, such as Bangla, Malayalam, Odiya, and Telugu, alongside metropolitan Indian accents and native accents from American, Australian, British, and Welsh English. The dataset is structured to meet key requirements for ASR development, including a variety of speakers, uniformity of content, and well-labeled data for training and testing models. To assess the accent detection capabilities of GPT-4o, we randomly selected 100 audio files from the AccentDB dataset. Each file was presented to the model with a prompt to identify the speaker's accent. The predictions made by GPT-4o were then compared to the ground truth labels to evaluate their performance.

The confusion matrix in Figure 15 highlights significant misclassifications, particularly with the Malayalam accent, which is frequently misclassified as Telugu. This misclassification suggests that the acoustic features of Malayalam and Telugu might be similar enough to confuse the model, indicating a need for more distinctive feature extraction and training data augmentation. Bangla and Telugu also exhibit substantial misclassification errors, particularly in Malayalam. This pattern suggests a broader challenge in differentiating between the phonetic characteristics of these languages, necessitating further refinement in the model's training process. The precision, recall, and F1-score metrics provide additional insights into the model's performance across different classes. The model demonstrates the highest precision for Odiya, indicating that when it predicts Odiya, it is often correct. However, the low recall for Odiya means that many Odiya instances are not being correctly identified. Malayalam shows a more balanced performance with relatively higher recall and F1-scores, suggesting that the model can correctly identify Malayalam instances more frequently. Both Bangla and Telugu have consistently low precision, recall, and F1-scores, indicating significant challenges in accurately detecting these accents. This demonstrates GPT-4o's limited ability to recognize and differentiate between various English accents, which is essential for enhancing the usability of ASR systems in multilingual and multicultural environments.

# Figure 15. Confusion Matrix (Top) and Performance Comparison (Bottom) for Accent Detection

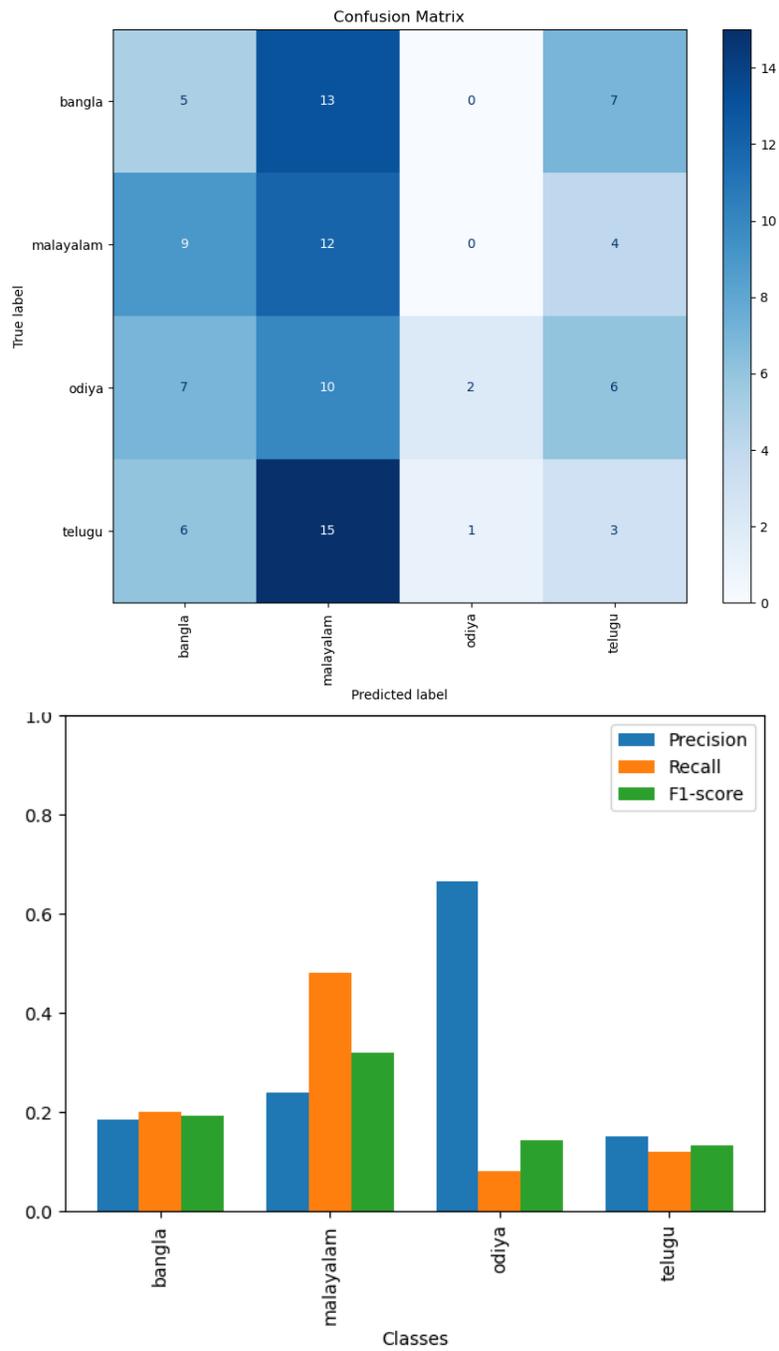

.

# 5. Multimodal Capacity of GPT-4o

The ability to integrate and interpret information from multiple modalities is crucial for developing advanced intelligent systems. Multimodal capacity refers to the capability of a model to understand and synthesize information from various sources such as text, images, and audio. This enables the model to generate more comprehensive and contextually enriched responses. The objective of assessing GPT-4o's multimodal capacity is to evaluate its performance across tasks that require the integration of different types of data.

## 5.1 Visual Question Answering

The Visual Question Answering (VQA) dataset is a multimodal benchmark that combines computer vision and NLP tasks. It consists of images paired with natural language questions related to the visual content[11]. The goal is to produce accurate natural language answers by comprehending the semantics of both the image and the question. For this evaluation, a subset of 100 image-question pairs was sampled from the dataset. GPT-4o was tasked with analyzing the provided image and the corresponding question and generating an appropriate answer chosen from a predefined list of possible answers. The model's generated answers were compared against the ground truth answers to assess its performance in this AI-complete task, which involves a wide range of sub-problems such as object detection, scene classification, and multimodal reasoning. The maximum accuracy was 0.36, as shown in Figure 16.

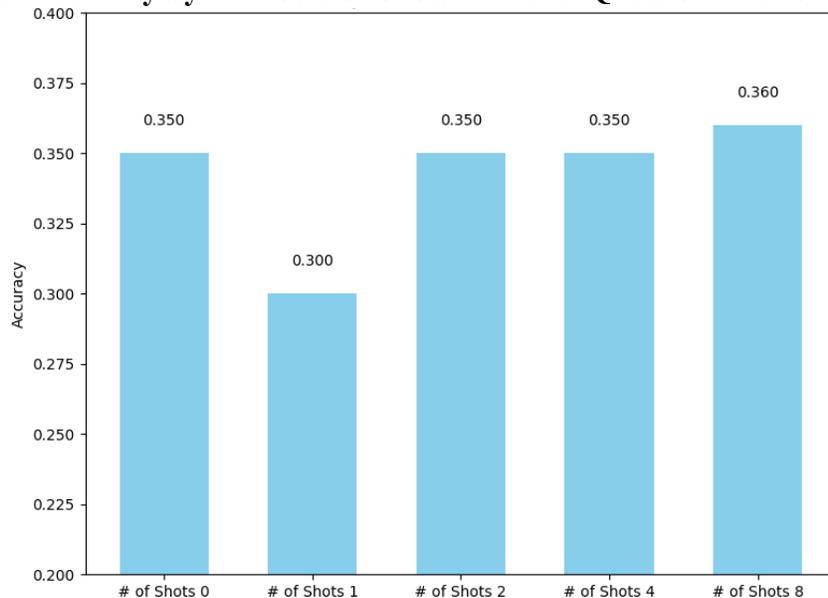

**Figure 16. Accuracy by Numbers of Shots in Visual Question Answering (VQA)**

The performance shows some variability with different shot numbers, peaking at 0.36 accuracy with eight shots. Interestingly, the model's performance decreases with just one example, suggesting that providing few examples in a task with many options may not always be beneficial. This decrease in performance could be due to the distribution of answers becoming skewed by the unrelated task, given the diverse possibilities in VQA.

---
[11] https://www.kaggle.com/datasets/bhavikardeshna/visual-question-answering-computer-vision-nlp

## 5.2 Vision-Language Capabilities

Vision-language (VL) capabilities represent a critical advancement in the development of AI models that can understand and interpret multimodal data, integrating both visual and linguistic information to perform complex tasks. The ability to combine these two types of data allows for a more nuanced understanding of content, which is essential for applications ranging from image captioning to more sophisticated tasks like explaining visual jokes or reasoning about events depicted in images.

To evaluate the vision-language capabilities of GPT-4o, we employed the MM-Vet benchmark (Yu et al., 2023). MM-Vet is designed to systematically assess large multimodal models (LMMs) on a variety of integrated tasks that require a combination of core VL capabilities, including recognition, optical character recognition (OCR), knowledge, language generation, spatial awareness, and math. This evaluation framework ensures comparison across diverse question types and answer styles and provides insights beyond simple performance rankings.

The MM-Vet benchmark includes tasks that necessitate the integration of these capabilities to solve complex problems. For instance, a task might involve recognizing objects in an image, understanding the spatial relationships between them, reading and interpreting text within the image, and generating a coherent textual response that incorporates external knowledge. The evaluation metrics employed by MM-Vet are based on an LLM-based evaluator that uses few-shot learning to provide scores for open-ended model outputs. This approach allows for consistent and comprehensive evaluation across different answer styles and question types. We compare the performance of GPT-4o with its predecessors in Table 14.

| Model | Recognition | OCR | Knowledge | Language Generation | Spatial Awareness | Math | Total |
|---|---|---|---|---|---|---|---|
| GPT-3.5 | 22.3 | 31.4 | 15.6 | 16.6 | 32.9 | 24.0 | 27.6 |
| GPT-4 | 34.3 | 66.3 | 25.6 | 36.6 | 60.6 | 72.0 | 48.1 |
| GPT-4V | 67.5 | 68.3 | 56.2 | 60.7 | 69.4 | 58.6 | 67.7 |
| GPT-4o | 80.6 | 87.5 | 75.7 | 77.1 | 84.7 | 88.1 | 83.9 |

Table 14. Performance Comparison of GPT Models on VL capabilities

The results from the MM-Vet benchmark highlight the advancements made with GPT-4o in VL capabilities compared to its predecessors. As summarized in Table 14, GPT-4o outperforms previous models across all evaluated metrics. GPT-4o's notable performance across all metrics highlights its advanced VL capabilities, setting a new benchmark for multimodal models. The high scores in knowledge, spatial awareness, and language generation tasks, in particular, highlight GPT-4o's ability to understand and produce contextually relevant responses based on visual inputs, making it versatile in various applications. The high performance noted is also consistent with that reported by Zhu et al., (2024). Table 15 provides examples of GPT-4o responses for several images and prompts.

| Input Image | Questions \| Assessed Capabilities | GPT-4o Responses |
|---|---|---|
| 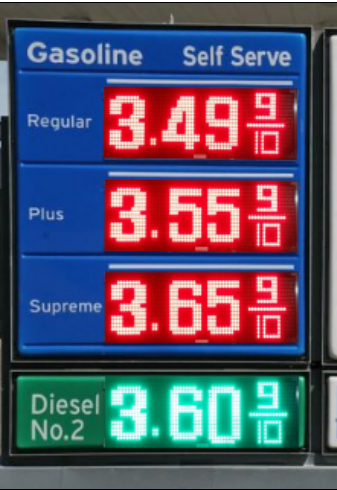 | a) Q: How many gallons of supreme gasoline can I get with $50? | GPT-4o: a) With $50, you can get approximately 13.70 gallons of supreme gasoline. |
| | b) Q: What is the price difference between diesel and regular gasoline? **Required capabilities: OCR, Math** | GPT-4o: The price difference between diesel and regular gasoline is $0.11 per gallon. |
| 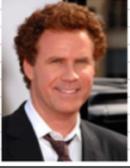 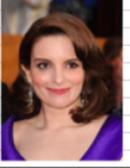 | a) Q: Based on the image, what is the total international awareness of Tina Fey? | GPT-4o: Based on the image, the total international awareness of Tina Fey is 35%. This percentage is located at the bottom row under the column labeled "Total" for Tina Fey. |
| | b) Q: Based on the image, in which country does Tina Fey have the highest level of total awareness? **Required capabilities: OCR, Spatial awareness** | GPT-4o: Based on the image, Tina Fey has the highest level of total awareness in Australia, with an awareness level of 62%. |
| 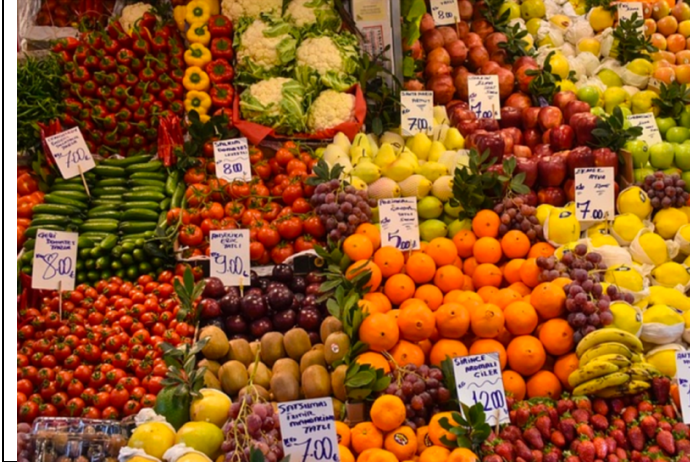 | a) Q: What is the price for tomatoes? | GPT-4o: The price for tomatoes is 8.00, as indicated by the sign placed above the pile of tomatoes in the image. |
| | b) Q: What fruit is the the right of plums? **Required capabilities: Recognition, Spatial awareness** | GPT-4o: The fruit to the right of the plums in the image is pears. |
| | a) Q: Which car is on the parking spot 33? | GPT-4o: The parking spot labeled 33 is empty. |
| 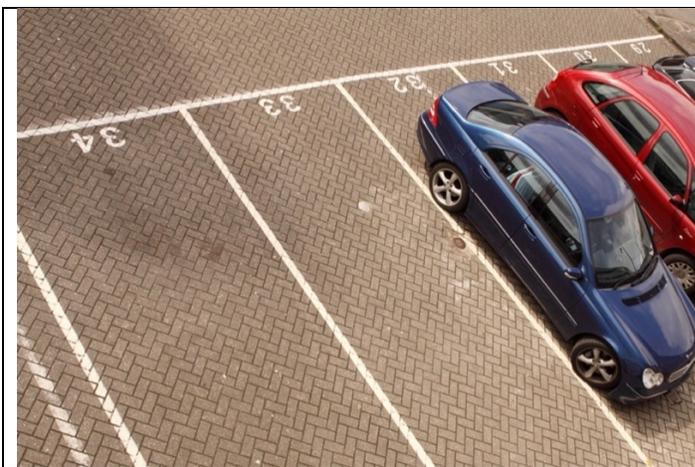 | b) Q: Which spot is the red car parking on?<br><br>**Required capabilities: OCR, Recognition, Spatial awareness** | GPT-4o: The red car is parked in spot number 30. |
| 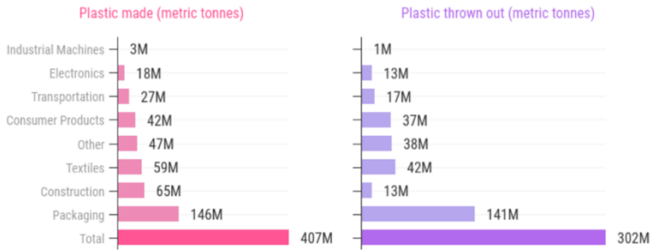 | a) Q: Which category did the worst job in plastic recycling, considering the amount of plastic made and thrown out? | GPT-4o: The category that did the worst job in plastic recycling, considering the amount of plastic made and thrown out, is "Packaging." In 2015, 146 million metric tonnes of plastic were made for packaging, and 141 million metric tonnes were thrown out. This shows that almost all the plastic made for packaging was discarded as waste. |
| | b) Q: What is the difference in metric tonnes between the amount of plastic produced and the amount discarded?<br><br>**Required capabilities: OCR, Spatial awareness** | The difference between the amount of plastic produced (407 million metric tonnes) and the amount discarded (302 million metric tonnes) is:<br><br>407 million metric tonnes – 302 million metric tonnes = 105 million metric tonnes<br><br>So, the difference is 105 million metric tonnes. |
| 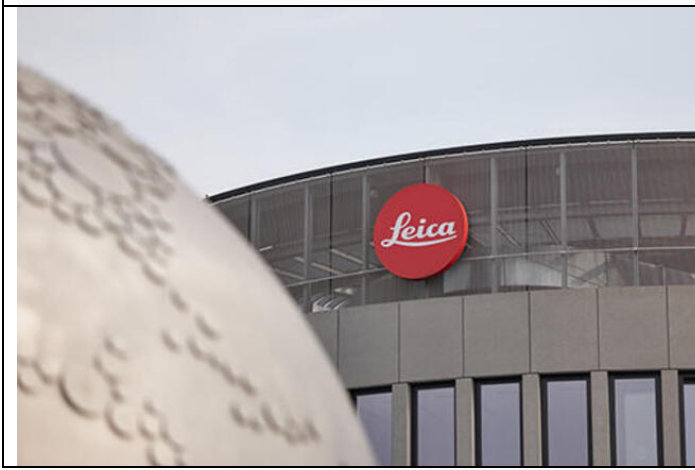 | Q: What does this factory make?<br><br>**Required capabilities: OCR** | GPT-4o: The factory in the image belongs to Leica, which is a company known for manufacturing cameras, lenses, and other optical equipment. |
| 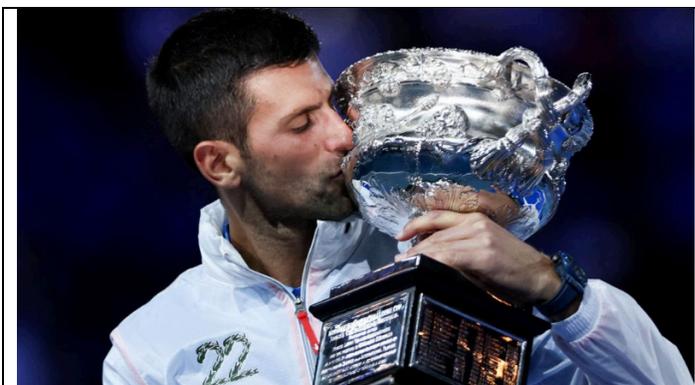 | Q: In which country was this photo taken?<br><br>**Required capabilities: Recognition, Knowledge** | GPT-4o: The photo shows Novak Djokovic kissing a trophy that appears to be the Australian Open trophy. Therefore, this photo was likely taken in Australia. |
| 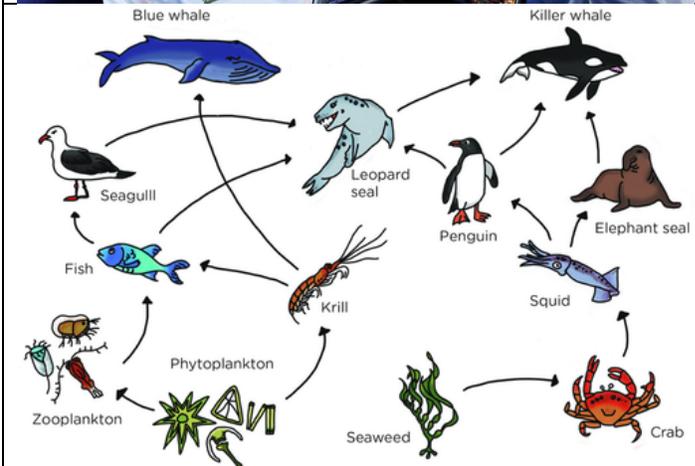 | Q: Which are producers in this food web?<br><br>**Required capabilities: OCR, Knowledge, Spatial Awareness** | GPT-4o: In this food web, the producers are:<br><br>Phytoplankton<br>Seaweed<br>These are the organisms that produce energy through photosynthesis and serve as the base of the food web. |
| 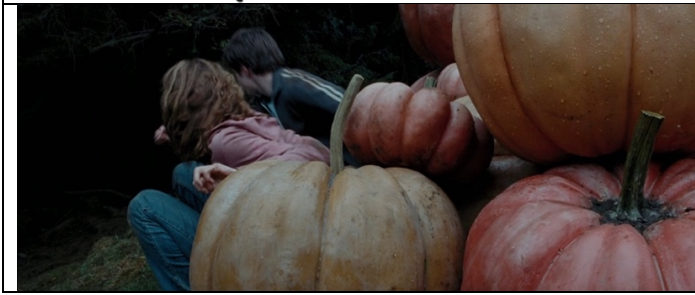 | Q: Is it summertime in the scene? Answer it and give the rationale.<br><br>**Required capabilities: Recognition, Knowledge, Language generation** | GPT-4o: No, it is not summertime in the scene. The presence of large pumpkins suggests that it is likely fall, as pumpkins are typically associated with the autumn harvest season. Additionally, the characters are wearing long-sleeved clothing, which is more common in cooler weather. |

| | Q: Can you explain this meme?<br><br>**Required capabilities: Recognition, OCR, Knowledge, Language generation** | GPT-4o: This meme humorously illustrates the common tendency to procrastinate. The text shows an intention to start a task at a specific time (8:00), but when the time is slightly missed (8:05), the person decides to delay the task further, waiting until the next hour (9:00) to start. The image of Kermit the Frog relaxing on a bed adds to the humor by depicting the relaxed attitude towards procrastination. The meme pokes fun at the irrational yet relatable habit of postponing tasks due to missing an arbitrary starting time by a few minutes. |
|---|---|---|
| 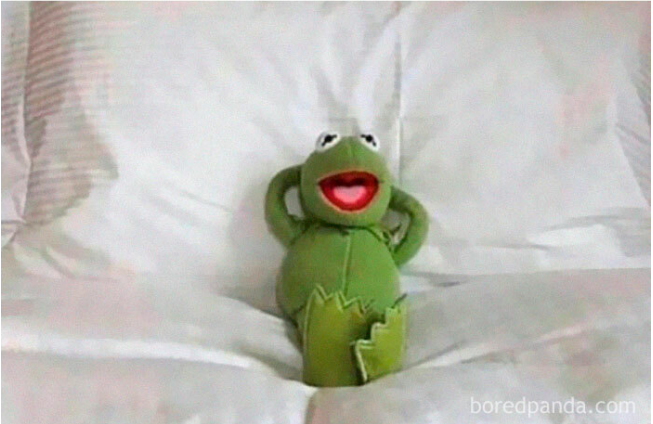 | | |
|  | Q: What is funny about this image?<br><br>**Required capabilities: Recognition, Knowledge, Language generation** | GPT-4o: This image is a popular meme that combines two unrelated photos to create a humorous juxtaposition.<br><br>    On the left side, there is an image of a woman (Taylor Armstrong from "The Real Housewives of Beverly Hills") angrily pointing and shouting, while another woman (Kyle Richards) tries to calm her down.<br><br>    On the right side, there is a photo of a white cat (known as "Smudge the Cat") sitting at a dinner table with a confused and displeased expression in front of a plate of vegetables.<br><br>The humor comes from the contrast between the woman's intense emotion and the cat's indifferent and bemused reaction, as if the cat is unbothered by the woman's outburst. This combination creates a relatable and funny depiction of exaggerated reactions and misunderstandings. |
| 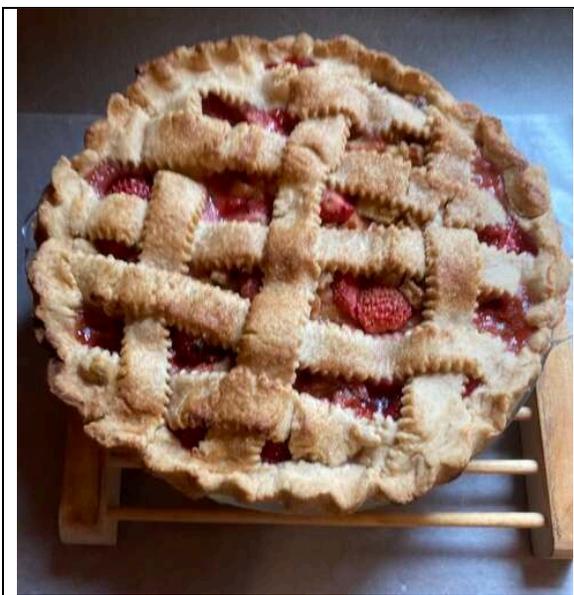 | Q: How to cook this dish?<br><br>**Required capabilities: Recognition, Knowledge, Language generation** | GPT-4o: This image shows a pie with a lattice crust. Here is a basic recipe to make a lattice-top pie, such as a strawberry or mixed berry pie:<br><br>Ingredients: For the Filling: 4 cups fresh strawberries or mixed berries, 3/4 cup granulated sugar, 1/4 cup cornstarch, 1 tablespoon lemon juice, 1 teaspoon vanilla extract, A pinch of salt For the Crust: 2 1/2 cups all-purpose flour, 1 teaspoon salt, 1 teaspoon sugar, 1 cup unsalted butter, chilled and cut into small pieces, 1/4 to 1/2 cup ice water<br><br>Instructions: In a large bowl, combine the flour, salt, and sugar. Add the chilled butter pieces and cut them into the flour mixture using a pastry blender or your fingers until the mixture resembles coarse crumbs. Gradually add ice water, a tablespoon at a time, mixing until the dough begins to come together. Do not overwork the dough. Divide the dough into two equal parts, shape them into disks, wrap in plastic wrap, and refrigerate for at least 1 hour. In a large bowl, combine the berries, sugar, cornstarch, lemon juice, vanilla extract, and salt. Stir gently to combine. On a lightly floured surface, roll out one disk of dough to fit a 9-inch pie pan. Place the rolled-out dough into the pie pan and trim the edges. Pour the berry mixture into the crust-lined pie pan. Roll out the second disk of dough and cut it into 1-inch-wide strips. Lay 4 to 5 strips over the filling, parallel to each other. Lay another 4 to 5 strips perpendicular to the first set, weaving them to create a lattice pattern. Trim and crimp the edges to seal. Preheat the oven to 375°F (190°C). Bake: Place the pie on a baking sheet to catch any drips and bake for 45-55 minutes, or until the crust is golden brown and the filling is bubbly. Cool: Let the pie cool completely before serving to allow the filling to set. Enjoy your homemade lattice-top pie! |
| 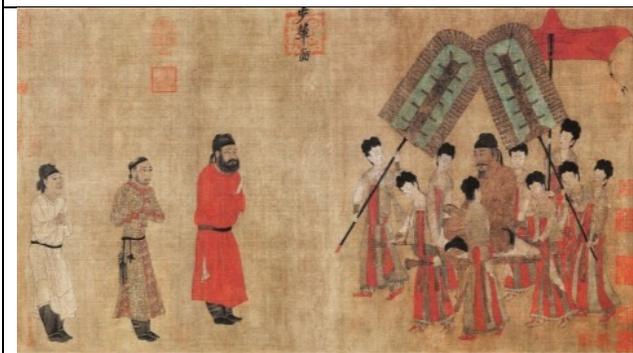 | Q: Can you give a short introduction to this painting?<br><br>**Required capabilities: Recognition, Knowledge, Language generation** | GPT-4o: This painting is a historical Chinese artwork titled "Emperor Taizong Receiving the Tibetan Envoy," created by Yan Liben during the Tang Dynasty. The piece illustrates a significant diplomatic event where Emperor Taizong of the Tang Dynasty meets a Tibetan envoy. This event highlights the political and cultural exchanges between the Tang Dynasty and Tibet. The painting is notable for its detailed depiction of the figures, the hierarchical arrangement, and the ceremonial aspects of the meeting. Yan Liben, a renowned court artist of the Tang Dynasty, is celebrated for his ability to capture the grandeur and solemnity of imperial events. This artwork not only serves as a historical record but also exemplifies the artistic sophistication of the Tang Dynasty. |

Table 15. Instances of GPT-4o Responses with Visual Inputs (incorrect answers highlighted in red)

# 6. Implications, Limitations, and Future Work

This section summarizes the key implications of our findings, acknowledges the limitations of the study, and outlines potential directions for future research.

## 6.1 Implications

The findings from this research have significant implications for the development and application of LLMs in various fields. GPT-4o's high performance in tasks like medical exam question answering and financial analysis suggests its potential utility in educational and professional training environments. The model's ability to integrate vision and language data effectively positions it as a valuable tool in fields requiring multimodal analysis, such as healthcare, finance, and customer service. The demonstrated proficiency in few-shot learning highlights the model's potential for applications where data is scarce or expensive. This could lead to more accessible AI-driven solutions in underrepresented languages and domains, offering inclusivity and broader application of AI technologies.

Moreover, the need to evaluate newer models on comprehensive and diverse sets of data and tasks is underscored by this research. The gap in robust and extensive evaluations has been a notable limitation in understanding the full capabilities and potential weaknesses of advanced models like GPT-4o. This calls for the development and adoption of more comprehensive benchmarks that can rigorously test models across a wider array of real-world scenarios. The findings also suggest implications for policy and regulatory frameworks. As AI models become increasingly integrated into critical sectors such as healthcare and finance, ensuring their reliability, transparency, and fairness becomes necessary (Hayawi & Shahriar, 2024). This necessitates continuous monitoring, rigorous testing, and the establishment of standards to guide the ethical deployment of AI technologies.

## 6.2 Limitations

Despite the promising results presented in this study, several limitations must be acknowledged. Firstly, the evaluation datasets used in various tasks, particularly in image and audio data, were relatively small and not exhaustive. This limited sample size may not fully capture the model's performance across all potential scenarios. While we aimed for a comprehensive evaluation across data types and multimodal (breadth), the categories within each are not exhaustive (depth). For example, we did not evaluate image and audio generation as it was beyond the scope of this study.

Moreover, qualitative or human judgment was not used as a criterion to assess performance. Incorporating human judgment is crucial for evaluating the practical usability and contextual accuracy of model outputs, as it provides insights that quantitative metrics alone may not reveal (Shahriar, 2022). The model also exhibited inconsistencies in handling ambiguous or complex inputs, as seen in the varying accuracy rates across different tasks. Furthermore, the few-shot learning approach, although beneficial in some contexts, showed limitations in tasks with a high degree of variability, such as VQA. The potential for overfitting to specific examples in these cases remains a concern. Additionally, the lack of real-time and longitudinal data evaluation poses a constraint on understanding the model's adaptability and robustness over time. For example, evaluating the model's performance in real-time applications, such as continuously monitoring driver drowsiness or detecting sudden changes in patient health through medical imaging, would provide valuable insights into its practical effectiveness and reliability under dynamic conditions.

## 6.3 Future Work

Building on the existing research, this paper highlights several avenues for future research directions. Expanding the evaluation datasets to include a more diverse and comprehensive range of tasks will provide a deeper understanding of the model's capabilities and limitations. Integrating real-time and longitudinal data assessments can offer insights into the model's adaptability and performance stability over extended periods. Further refinement of the few-shot learning techniques is essential, especially for tasks with high variability. Exploring advanced prompting strategies and incorporating more contextual understanding (Sivarajkumar et al., 2024) could enhance performance in these areas. It is thus important to also investigate the impact of prompt quality on model performance. Additionally, understanding the reasons behind the model's low performance and conducting thorough error analysis are crucial. This involves examining how and why the model failed in specific tasks to inform targeted training and fine-tuning efforts. Such analysis will provide valuable insights into the model's limitations and guide improvements to enhance its utility in nuanced language understanding tasks.

Future work should also prioritize creating and adopting new, comprehensive benchmarks that evaluate models across diverse tasks and datasets, addressing the current gap in robust model evaluation. This approach will ensure a holistic understanding of the model's performance, guiding improvements and encouraging the development of more reliable AI systems. The current multimodal evaluation only investigated image and text inputs, highlighting the necessity to explore other inputs and their combinations. For instance, incorporating audio, image, and text together could significantly contribute to cross-domain applications and arts (Shahriar & Al Roken, 2022), enhancing the model's utility in various fields. Lastly, incorporating qualitative assessments and human judgment in the evaluation process will provide a more nuanced understanding of the model's practical applicability and contextual performance. This can help identify areas where the model performs well in real-world scenarios and where it may require further enhancement.